\documentclass[conference]{IEEEtran}
\IEEEoverridecommandlockouts
\usepackage{cite}
\usepackage{amsmath,amssymb,amsfonts}
\usepackage{algorithmic}
\usepackage{hyperref}
\usepackage{graphicx}
\usepackage{textcomp}
\usepackage{xcolor}
\usepackage{multirow}
\usepackage{graphicx}
\usepackage{adjustbox}
\usepackage{booktabs}   
\usepackage{multirow}   
\usepackage{makecell}   
\usepackage{array}      
\usepackage{pifont}

\hypersetup{
    colorlinks=true,    
    linkcolor=black,    
    citecolor=blue,    
    urlcolor=black      
}
\usepackage[all]{hypcap}

\def\BibTeX{{\rm B\kern-.05em{\sc i\kern-.025em b}\kern-.08em
    T\kern-.1667em\lower.7ex\hbox{E}\kern-.125emX}}
\begin{document}

\title{Image-Conditioned Instance Prompt Network for Referring Remote Sensing Image Segmentation}

\author{
\IEEEauthorblockN{
    Biaoyu Ren$^{1\dagger}$, 
    Qingsheng Wang$^{1\dagger}$, 
    Cun Xu$^{1}$,
    Dingkang Yang$^{2*}$,
    Wenxuan Wang$^{1,3*}$
}
\IEEEauthorblockA{
    $^1$School of Computer Science, Northwestern Polytechnical University, Xi'an, China \\
    $^2$College of Intelligent Robotics and Advanced Manufacturing, Fudan University, Shanghai, China \\
    $^3$Shenzhen Research Institute of Northwestern Polytechnical University, Shenzhen, China \\
    \{renby, wqshmzh, xucun\}@mail.nwpu.edu.cn, \\
    dkyang20@fudan.edu.cn,  wxwang@nwpu.edu.cn
}
\thanks{$^\dagger$Equal contribution. $^*$Corresponding authors.}
\thanks{Code available at: \url{https://github.com/Ren-by/ICIPNet}.}
}
\maketitle


\begin{abstract}
Referring Remote Sensing Image Segmentation (RRSIS) is a situated, task-driven cross-modal task related to the embodied perception paradigm, requiring models to align visual-spatial features with linguistic intentions for precise target perception. Recent research has focused on refining the granularity of textual features and optimizing image-text feature fusion to better guide target feature representations. However, insufficient descriptive granularity and sensitivity to semantic shifts can cause bottlenecks in cross-modal feature fusion. To address these issues, we propose the Image-Conditioned Instance Prompt Network (ICIPNet) with Bilateral Information Fusion, which is designed to alleviate bottlenecks in cross-modal feature fusion. ICIPNet introduces an Image-Conditioned Instance Prompt (ICIP) module to generate self-adaptive visual and semantic representations without external knowledge. The Bilateral Information Fusion (BIF) module enhances feature fusion along the token and channel dimensions. Experiments demonstrate that the proposed ICIPNet outperforms existing RRSIS models.
\end{abstract}


\section{Introduction}
\label{sec:intro}
Referring Remote Sensing Image Segmentation (RRSIS) aims to segment specified objects in remote sensing images based on natural language expressions. It is an important cross-modal task related to the embodied perception paradigm, with broad applications in ground object recognition, urban management, and environmental monitoring. In contrast to traditional image segmentation methods, the RRSIS requires a comprehensive understanding of object attributes and geographic details, which makes the task more challenging. 

\begin{figure}[!t]
\vspace{-0.5mm}    
\centering
\includegraphics[width=\linewidth]{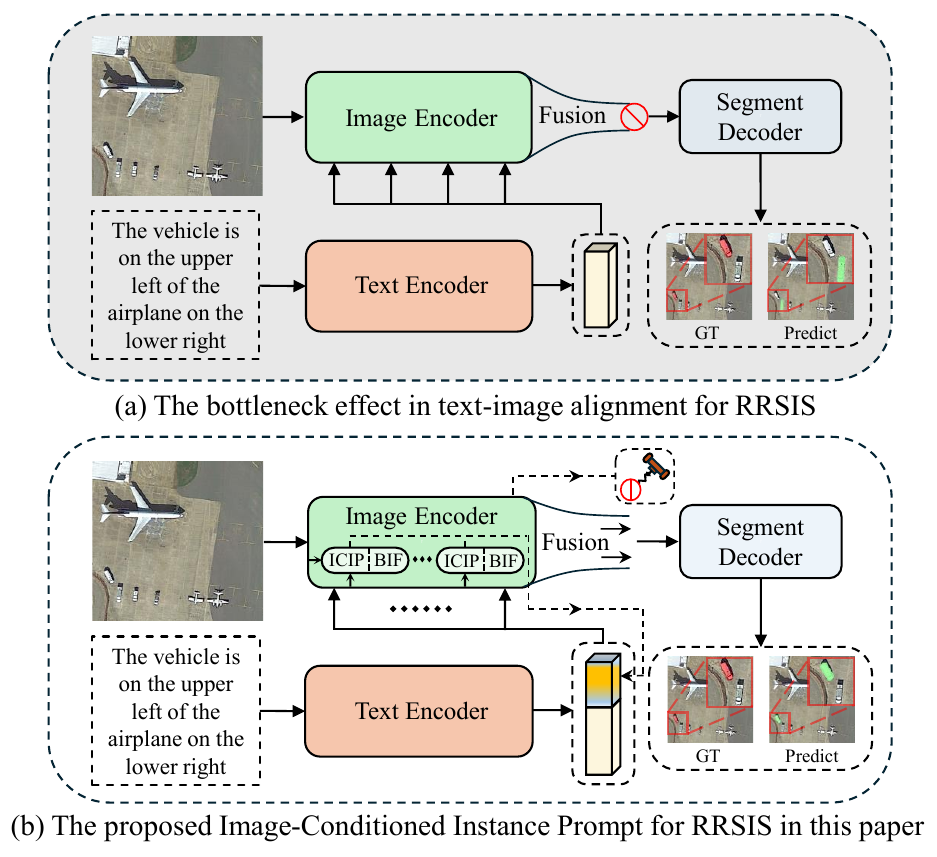}   
\vspace{-6mm}
\caption{The motivation of our proposed model. Figure (a) illustrates how previous models fuse vision–language features by leveraging a fixed language distribution. Figure (b) describes the proposed ICIPNet, where the semantic prompts no longer rely on a fixed and uneven language distribution, but are influenced by the dynamic stretching and redistribution of visual content. This adaptively adjusts the semantic features across layers and enables the model to dynamically focus on key regions in remote sensing images.}
\vspace{-4mm}
\label{figure_1_1}
\end{figure}

In the literature, one of the research categories on referring remote sensing image segmentation \cite{liu2024rotated,lei2024exploring,yao2025remotesam,zhang2025referring,shi2025multimodal,yuan2024rrsis} includes directly integrating textual features into visual features at intermediate layers of the visual encoder, followed by predicting the segmentation masks using a lightweight segmentation decoder. As shown in \hyperref[figure_1_1]{Fig.~1} (a), the textual features of the natural language expressions are fused with the visual features using a pixel-level attention mechanism, which guides the extraction of the target features. However, these works are sensitive to insufficient language granularity and semantic shifts\cite{zhao-etal-2024-representation}. To address this issue, we revisit the conventional alignment approach and propose to leverage semantic self-adaptation from visual contexts in the input images. As illustrated in \hyperref[figure_1_1]{Fig.~1} (b), ICIPNet enables the semantic prompts to be influenced by the dynamic feature distributions of the
\noindent visual contents across input images, instead of relying on a fixed and uneven language distribution from the language backbone, which in turn adjusts the attention regions in the input images across intermediate layers of the image encoder, thereby ensuring semantic consistency across hierarchical levels.

Conventional cross attention captures irrelevant visual patch features and language features, leading to high sensitivity to target backgrounds in the input images and high-frequency co-occurring words (e.g., "object", "area") in the natural language expressions. To address this, we propose Bilateral Information Fusion (BIF) module to dynamically suppress attention noise between visual and language modalities and better reconstruct the encoded features. Specifically, query and key features are split to conduct separate attention operations to suppress attention noise and the feature reconstruction after the attention is conducted by another bilateral fusion strategy, which is Multi-modal Feature Reconstruction (MFR) that implicitly models and applies nonlinear transformations along the token and channel dimensions.

\begin{figure*}[!t]
\vspace{-0.5mm}    
\centering
\includegraphics[width=\linewidth]{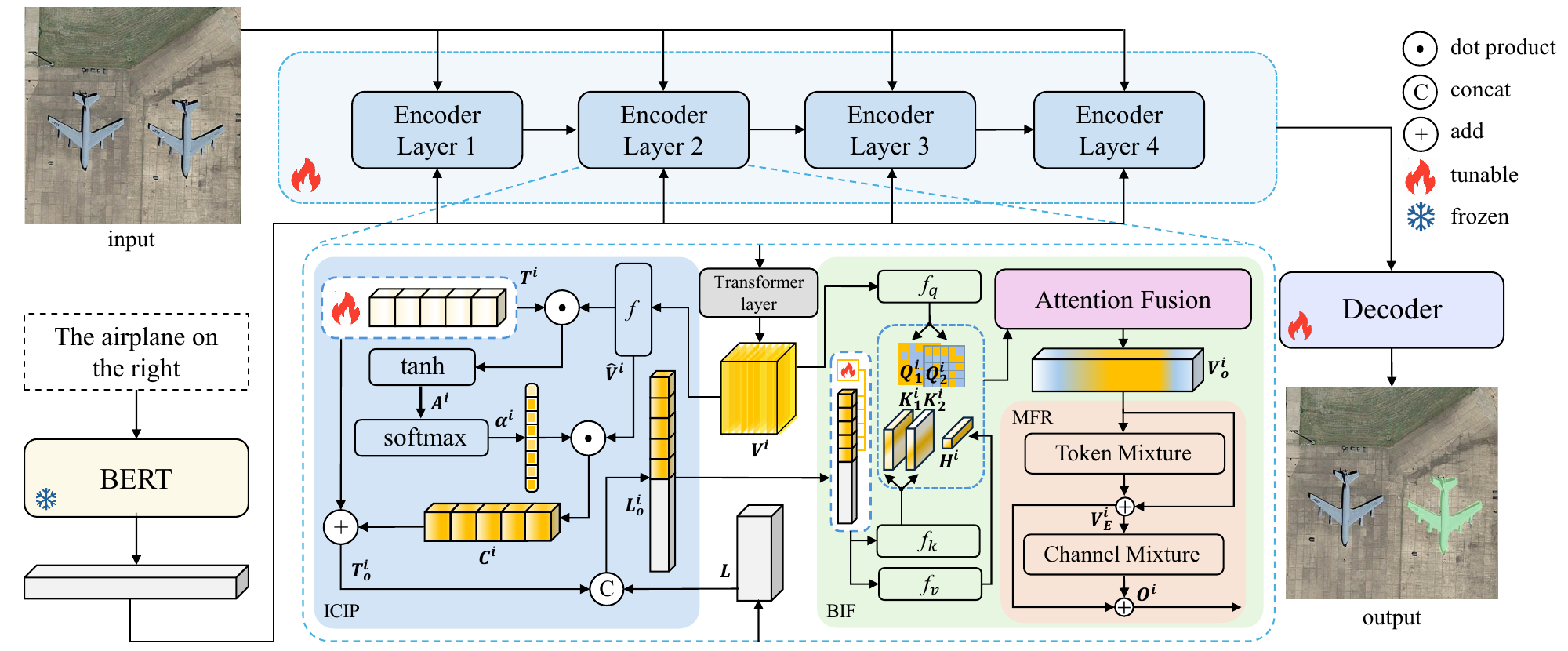}   
\vspace{-6mm}
\caption{Overall pipeline of the proposed ICIPNet. Our model is based on a hierarchical language-aware visual encoding via a pre-trained Swin Transformer. At the $i$-th layer, the input image is encoded into the input image visual feature matrix $V^i, i \in \{1,2,3,4\}$. In the Image-Conditioned Instance Prompt (\hyperref[sec:ICIP]{ICIP}) module introduced in (Sec.~\ref{sec:ICIP}), $V^i$ is used to adaptively generate $T_o^i$, which is then concatenated with the language features matrix $L$. Subsequently, the Bilateral Information Fusion module (\hyperref[sec:BIF]{BIF})  (Sec.~\ref{sec:BIF}) fuses $L_o^i$ with $V_i$ to produce enhanced multimodal fused features $O$. Finally, a standard segmentation decoder generates the final pixel-level segmentation output.}
\vspace{-4mm}
\label{figure_3_1}
\end{figure*}

The main contributions of this paper are summarized as follows:
\begin{itemize}
  \item We propose an ICIP module that learns image-conditioned instance prompts to extract flexible vision-language features varying across different input images.
  \item We design a BIF module to refine the visual features of input images by capturing critical vision-language features using bilateral cross attentions and MFR.
  \item We conduct extensive experiments on the RefSegRS and RRSIS-D datasets, achieving state-of-the-art (SOTA) performance.
\end{itemize}

\section{Related Work}
\label{section_2}
\subsection{Referring Remote Sensing Image Segmentation}
Remote sensing referring image segmentation (RRSIS) focuses on pixel-level localization of target objects conditioned on associated natural language expressions. Recent methods primarily concentrate on enhancing language granularity and strengthening vision–language alignment to improve segmentation performance \cite{liu2024rotated,lei2024exploring,yao2025remotesam,zhang2025referring,shi2025multimodal,yuan2024rrsis}. Despite these advancements, existing approaches still face challenges inherent to remote sensing images, including large-scale variations\cite{ma2023sacanet}, complex background clutter, and semantic discrepancies between visual content and textual descriptions.

\subsection{Instance-level Visual Prompt Tuning}
Instance-level visual prompt tuning dynamically generates instance-specific prompts for each input image \cite{xiao2025visual,lu2025task,yao2025bi,xie2024pa,zhou2022cocoop}, aiming to specifically capture fine-grained features of different input images to achieve flexibility. While this approach can enhance object recognition to some extent, single-modality prompts are limited in modeling cross-modal correlations and struggle to maintain the spatial consistency of targets across multi-scale features. To address these limitations, our method is designed to overcome the modality constraints and spatial restrictions inherent in conventional prompts, enabling more flexible visual feature perception.

\section{Methodology}

This section introduces the proposed ICIPNet. As shown in  \hyperref[figure_3_1]{Fig.~2}, given an input RGB image 
\(
\boldsymbol{I} \in \mathbb{R}^{H \times W \times 3}
\) 
and a natural language expression 
\(
E = \{\omega_j\}_{j=1}^{N}
\), 
where \(H\) and \(W\) denote the height and width of $\boldsymbol{I}$, respectively, $\omega_j$ denotes the $j$-th word, and \(N\) represents the sentence length. The input natural language expressions in a batch are encoded into language feature matrix 
\(
\boldsymbol{L} \in \mathbb{R}^{B \times N \times C}
\) 
via a pre-trained BERT\cite{devlin-etal-2019-bert}, where \(B\) is the batch size and \(C\) is the number of channels of each word embedding. Hierarchical visual feature matrix 
\(
\boldsymbol{V}^{i} \in \mathbb{R}^{B \times P_i \times C_i}, i \in \{1,2,3,4\}
\) 
outputted from the $i$-th pre-trained Swin Transformer \cite{liu2021Swin} (ST) layer is extracted from $\boldsymbol{I}$, where \(P_i\) and \(C_i\) denote the number of image patches and the number of channels, respectively. The ICIP and BIF modules perform cross-modal modeling on the visual features \(V^i\) at each ST layer to extract more discriminative multi-modal features. Finally, a lightweight segmentation decoder generates pixel-level segmentation masks. Section~\ref{sec:ICIP} provides a detailed description of the ICIP module, while Section~\ref{sec:BIF} describes the BIF module.

\subsection{Image-Conditioned Instance Prompt}
\label{sec:ICIP}

We design the Image-Conditioned Instance Prompt (ICIP) module to adaptively generate language features conditioned on the visual contexts in different input images. Specifically, 
\noindent we perform attention-based interaction modeling between randomly initialized visual prompts and image patch features and adaptively aggregate visual contexts from the input images to form context-aware instance-level visual prompts.

The randomly initialized visual prompts at the $i$-th ST layer are placed in a matrix, which is:
\begin{equation}\label{eq:Ti}
\boldsymbol{T}^{i}
=
\{ \boldsymbol{t}_{1}^{i}, \boldsymbol{t}_{2}^{i}, \dots, \boldsymbol{t}_{M_i}^{i} \}
\in \mathbb{R}^{B \times M_i \times D}
\end{equation}
where $B$ denotes the batch size, $M_i$ is the manually defined number of visual prompt vectors, $\boldsymbol{t}_{M_i}^i\in\mathbb{R}^{B\times C_i}$ is a visual prompt matrix containing $B$ copied visual prompt vectors with $C_i$ channels, and $P_i$ denotes the number of image patches at the $i$-th ST layer.

First, we map $\boldsymbol{V}^{i}$ from $C_i$ to $D$ channels via an MLP $f$ to acquire $\hat{\boldsymbol{V}}^{i}=f(\boldsymbol{V}^{i})\in \mathbb{R}^{B \times P_i \times D}$.
In order to adaptively perceive the variations of target regions in the input images, we compute the similarity matrix $\boldsymbol{A}^{i}$ between $\hat{\boldsymbol{V}}^{i}$ and $\boldsymbol{T}^{i}$:
\begin{equation}\label{eq:Ai}
\boldsymbol{A}^{i}
=
\tanh
\left(
\hat{\boldsymbol{V}}^{i} (\boldsymbol{T}^{i})^{\top}
\right)
\in
\mathbb{R}^{B \times P_i \times M_i}
\end{equation}
where $\tanh()$ denotes the hyperbolic tangent activation function. The attention weight matrix $\boldsymbol{\alpha^{i}}\in\mathbb{R}^{B \times P_i \times M_i}$ between $\hat{\boldsymbol{V}}^{i}$ and $\boldsymbol{T}^{i}$ is then computed as:
\begin{equation}
\label{eq:alpha}
\boldsymbol\alpha^{i}_{b,p,m}
=
\frac{
\exp\left( \boldsymbol{A}^{i}_{b,p,m} \right)
}{
\sum_{p'=1}^{P_i}
\exp\left( \boldsymbol{A}^{i}_{b,p',m} \right)
}
\end{equation}
where $b \in \{1, \dots, B\}$ denotes the batch index, 
$p \in \{1, \dots, P_i\}$ denotes the index of the image patch at the $i$-th layer, 
$p' \in \{1, \dots, P_i\}$ denotes the image patch index used for summation over all patches in the softmax computation, 
and $m \in \{1, \dots, M_i\}$ denotes the index of the learnable instance prompts at the $i$-th layer.

Next, the image-conditioned instance prompts matrix $\boldsymbol{T}_o^{i}\in\mathbb{R}^{B \times M_i \times D}$ is preliminarily computed by residually aggregating the image patch features in $\hat{\boldsymbol{V}}^{i}$ according to the above attention weight matrix $\boldsymbol{\alpha^{i}}$:
\begin{equation}
\label{eq:Ci}
\boldsymbol{T}_o^{i}
=
(\boldsymbol{\alpha}^{i})^{\top}
\hat{\boldsymbol{V}}^{i} + \boldsymbol{T}^i
\end{equation}

The final image-conditioned instance prompts not only perceive visual contexts in the input images, but also capture language features of $E$. We concatenate $\boldsymbol{T}_o^{i}$ and $\boldsymbol{L}$ along the token dimension, resulting in the image-conditioned instance prompt matrix $\boldsymbol{L}_{o}^{i}=[\boldsymbol{T}_{o}^{i};\boldsymbol{L}]\in\mathbb{R}^{B \times (M_i + N) \times D}$ that incorporates both visual contextual features and language features.

\subsection{Bilateral Information Fusion}
\label{sec:BIF}
In this section, we provide a detailed description of the BIF module, as illustrated in  \hyperref[figure_3_1]{Fig.~2}. The module first performs dynamic feature modulation between $\boldsymbol{V}^i$ and  $\boldsymbol{L}_o^i$ at the $i$-th ST layer. Then, the Multi-Modal Feature Reconstruction (MFR) module reconstructs the modulated features and performs semantic disentanglement. These modules enable dynamic optimization and structured aggregation of cross-modal features, providing more discriminative and robust representations for vision-language understanding in complex remote sensing scenarios.

\subsubsection{Dynamic Feature Modulation} Following the work of \cite{ye2025differential}, we adaptively and linearly split the projected $\boldsymbol{L}_o^i$ and $\boldsymbol{V}^i$ into two sets to enable the model to dynamically suppress attention noise during the bilateral cross attention operations. The bilateral attention responses of two feature sets highlight semantically significant regions, thereby enhancing the recognition of critical areas.

First, $\boldsymbol{\mathit{V}}^i$ and $\boldsymbol{\mathit{L}}_o^i$ are linearly projected through a set of linear layers $f_q$, $f_k$, and $f_v$ to acquire query, key, and value matrices $\boldsymbol{Q}^i$, $\boldsymbol{K}^i$, and $\boldsymbol{H}^i$:
\begin{equation}
\boldsymbol{\mathit{Q}}^i = f_q(\boldsymbol{\mathit{V}}^i), \quad 
\boldsymbol{\mathit{K}}^i = f_k(\boldsymbol{\mathit{L}}_o^i), \quad 
\boldsymbol{\mathit{H}}^i = f_v(\boldsymbol{\mathit{L}}_o^i)
\end{equation}
where the shapes of $\boldsymbol{\mathit{Q}}^i$, $\boldsymbol{\mathit{K}}^i$, and $\boldsymbol{\mathit{H}}^i$ are all $B \times M_i \times 2D$. The two matrices $\boldsymbol{\mathit{Q}}^i$ and $\boldsymbol{\mathit{K}}^i$ are split along the channel dimension into two sets $\{\boldsymbol{Q}^i_1,\boldsymbol{K}^i_1\}$ and $\{\boldsymbol{Q}^i_2,\boldsymbol{K}^i_2\}$:
\begin{equation}
\begin{split}
\boldsymbol{\mathit{Q}}^i_1, \boldsymbol{\mathit{Q}}^i_2 &= \text{split}(\boldsymbol{\mathit{Q}}^i, \text{dim}=-1), \\
\boldsymbol{\mathit{K}}^i_1, \boldsymbol{\mathit{K}}^i_2 &= \text{split}(\boldsymbol{\mathit{K}}^i, \text{dim}=-1)
\end{split}
\end{equation}
where $\text{split}(\boldsymbol{Z}, \text{dim}=-1)$ denotes evenly dividing $\boldsymbol{Z}$ along the last dimension. To adaptively balance the contributions of different subspaces in cross-modal feature interactions, a learnable weighting coefficient $\boldsymbol{\mathit{\lambda}}_v^i$ is introduced, which is computed via the dot product of four Gaussian initialized and learnable parameters $\boldsymbol{\mathit{\lambda}}_{q1}^i$, $\boldsymbol{\mathit{\lambda}}_{k1}^i$, $\boldsymbol{\mathit{\lambda}}_{q2}^i$, and $\boldsymbol{\mathit{\lambda}}_{k2}^i$, whose shapes are $1\times D$:
\begin{equation}
\boldsymbol{\mathit{\lambda}}_v^i = \exp(\boldsymbol{\mathit{\lambda}}_{q1}^i \cdot \boldsymbol{\mathit{\lambda}}_{k1}^{i\ \top}) - \exp(\boldsymbol{\mathit{\lambda}}_{q2}^i \cdot \boldsymbol{\mathit{\lambda}}_{k2}^{i\ \top}) + \boldsymbol{\mathit{\lambda}}^i
\end{equation}
where $\boldsymbol{\mathit{\lambda}}_{q1}^i$ and $\boldsymbol{\mathit{\lambda}}_{k1}^i$ adjust the correlation of $\boldsymbol{\mathit{Q}}_1^i$ and $\boldsymbol{\mathit{K}}_1^i$ in the first subspace, $\boldsymbol{\mathit{\lambda}}_{q2}^i$ and $\boldsymbol{\mathit{\lambda}}_{k2}^i$ adjust the correlation of $\boldsymbol{\mathit{Q}}_2^i$ and $\boldsymbol{\mathit{K}}_2^i$ in the second subspace, and $\boldsymbol{\mathit{\lambda}}^i$ is a layer-wise dynamic bias term. The correlation matrix $\boldsymbol{F}^i$ of the two subspaces is then acquired as follows:
\begin{equation}
\boldsymbol{\mathit{F}}^i = \frac{1}{\sqrt{G_i}} (\boldsymbol{\mathit{Q}}_1^i \boldsymbol{\mathit{K}}_1^{i\ \top}) - \frac{\boldsymbol{\mathit{\lambda}}_v}{\sqrt{G_i}} (\boldsymbol{\mathit{Q}}_2^i \boldsymbol{\mathit{K}}_2^{i\ \top})
\end{equation}
where $G_i$ is the number of feature channels.

Finally, the original features are weighted $\boldsymbol{F}^i$:
\begin{equation}
\boldsymbol{\mathit{V}}_o^i = \boldsymbol{\mathit{F}}^i \boldsymbol{\mathit{H}}^i
\end{equation}

\subsubsection{Bilateral Feature Reconstruction}
The Bilateral Feature Reconstruction (BFR) module performs implicit modeling of feature distributions, enabling the reconstruction and aggregation of feature representations. We first perform token mixture, which operates along the token dimension, to model global contextual relationships and cross-position semantic correlations, thereby improving the overall feature consistencies across different tokens. Then, a channel mixture operation is performed for each token to capture dependencies across channels, facilitating fine-grained feature interactions and enhancing discriminative capacity.

\begin{table*}[htbp]
\label{tab:refsegrs}
\centering
\caption{Results of referring image segmentation with different models on the \textbf{RefSegRS} dataset. The best results are in bold.}
\label{tab:refsegrs}
\resizebox{\textwidth}{!}{%
\begin{tabular}{l | c | >{\centering\arraybackslash}p{1.2cm} | >{\centering\arraybackslash}p{1.2cm} 
| c|c | c|c | c|c | c|c | c|c | c|c | c|c}
\toprule
\multirow{2}{*}{Method} & \multirow{2}{*}{Publication} & \multirow{2}{*}{\makecell[c]{Visual \\ Encoder}} & \multirow{2}{*}{\makecell[c]{Text \\ Encoder}} &
\multicolumn{2}{c|}{P@0.5} & \multicolumn{2}{c|}{P@0.6} & \multicolumn{2}{c|}{P@0.7} &
\multicolumn{2}{c|}{P@0.8} & \multicolumn{2}{c|}{P@0.9} & \multicolumn{2}{c|}{mIoU} & \multicolumn{2}{c}{oIoU} \\
\cline{5-18}
 & & & & 
 \raisebox{-0.2em}{Val} & \raisebox{-0.2em}{Test} &
 \raisebox{-0.2em}{Val} & \raisebox{-0.2em}{Test} &
 \raisebox{-0.2em}{Val} & \raisebox{-0.2em}{Test} &
 \raisebox{-0.2em}{Val} & \raisebox{-0.2em}{Test} &
 \raisebox{-0.2em}{Val} & \raisebox{-0.2em}{Test} &
 \raisebox{-0.2em}{Val} & \raisebox{-0.2em}{Test} &
 \raisebox{-0.2em}{Val} & \raisebox{-0.2em}{Test} \\
\midrule
RMSIN & CVPR'2024 & Swin-B & BERT & 93.04 & 75.45 & 88.40 & 63.73 & 77.26 & 38.14 & 33.87 & 14.14 & 7.42 & 2.97 & 74.06 & 60.23 & 81.58 & 74.64 \\
MAFN & GRSL'2025 & Swin-B & BERT & 95.59 & 78.32 & 91.18 & 67.42 & 82.37 & 42.93 & 45.48 & 16.73 & 8.35 & 3.03 & 76.53 & 62.36 & 82.04 & 74.26 \\
FIANet & TGRS’2025 & Swin-B & BERT & 94.90 & 78.22 & 92.81 & 68.17 & 84.22 & 40.22 & 47.33 & 18.98 & \textbf{9.98} & 3.08 & 77.18 & 62.89 & 82.51 & 74.94 \\
RemoteSAM & MM'2025 & Swin-B & BERT & \textbf{96.29} & 79.20 & \textbf{93.97} & 67.20 & 84.69 & 43.53 & 45.71 & 17.45 & 8.35 & 3.14 & 77.31 & 62.90 & 82.69 & 74.77 \\
Ours & —— & Swin-B & BERT & 96.23 & \textbf{82.39} & 93.93 & \textbf{73.77} & \textbf{87.40} & \textbf{57.47} & \textbf{51.81} & \textbf{22.84} & 9.54 & \textbf{4.64} & \textbf{78.25} & \textbf{66.08} & \textbf{83.78} & \textbf{76.55} \\

\bottomrule
\end{tabular}%
}
\end{table*}

\begin{table*}[htbp]
\caption{Results of referring image segmentation with different models on the \textbf{RRSIS-D} dataset. The best results are in bold.}
\label{tab:rrsis-d}
\centering
\resizebox{\textwidth}{!}{%
\begin{tabular}{
l | c | 
>{\centering\arraybackslash}p{1.2cm} | 
>{\centering\arraybackslash}p{1.2cm} |
c|c | c|c | c|c | c|c | c|c | c|c | c|c }
\toprule
\multirow{2}{*}{Method} &
\multirow{2}{*}{Publication} &
\multirow{2}{*}{\makecell[c]{Visual \\ Encoder}} &
\multirow{2}{*}{\makecell[c]{Text \\ Encoder}} &
\multicolumn{2}{c|}{P@0.5} &
\multicolumn{2}{c|}{P@0.6} &
\multicolumn{2}{c|}{P@0.7} &
\multicolumn{2}{c|}{P@0.8} &
\multicolumn{2}{c|}{P@0.9} &
\multicolumn{2}{c|}{mIoU} &
\multicolumn{2}{c}{oIoU} \\
\cline{5-18}
 & & & & 
 \raisebox{-0.2em}{Val} & \raisebox{-0.2em}{Test} &
 \raisebox{-0.2em}{Val} & \raisebox{-0.2em}{Test} &
 \raisebox{-0.2em}{Val} & \raisebox{-0.2em}{Test} &
 \raisebox{-0.2em}{Val} & \raisebox{-0.2em}{Test} &
 \raisebox{-0.2em}{Val} & \raisebox{-0.2em}{Test} &
 \raisebox{-0.2em}{Val} & \raisebox{-0.2em}{Test} &
 \raisebox{-0.2em}{Val} & \raisebox{-0.2em}{Test} \\
\midrule
RMSIN & CVPR'2024 & Swin-B & BERT &
72.70 & 71.70 &
64.89 & 65.10 &
54.43 & 53.92 &
42.30 & 41.25 &
22.82 & 22.46 &
63.14 & 62.37 &
76.31 & 75.91 \\
MAFN & GRSL'2025 & Swin-B & BERT &
74.63 & 73.35 &
67.22 & 67.20 &
56.47 & \textbf{56.28} &
43.43 & 42.61 &
24.75 & 23.62 &
64.08 & 63.21 &
77.23 & 76.14 \\
FIANet & TGRS’2025 & Swin-B & BERT &
73.74 & 73.20 &
67.64 & 66.79 &
55.92 & 55.73 &
43.62 & 41.11 &
25.23 & 24.10 &
63.87 & 63.01 &
76.95 & 76.35 \\
RemoteSAM & MM'2025 & Swin-B & BERT &
73.60 & 73.56 &
66.59 & 65.98 &
\textbf{57.47} & 54.48 &
43.77 & 42.47 &
\textbf{25.75} & 24.39 &
64.11 & 63.31 &
77.04 & 76.41 \\
Ours & —— & Swin-B & BERT &
\textbf{75.17} & \textbf{73.84} &
\textbf{67.83} & \textbf{67.56} &
56.60 & 56.11 &
\textbf{44.06} & \textbf{43.20} &
24.99 & \textbf{24.63} &
\textbf{65.19} & \textbf{63.93} &
\textbf{77.69} & \textbf{76.87} \\
\bottomrule
\end{tabular}%
}
\vspace{-3mm}
\end{table*}
The token mixture is formulated as follows:
\begin{equation}
\boldsymbol{O}_{\mathrm{TM}}^i = R_2 \big( P_{\mathrm{token}}( R_1(\boldsymbol{V}_o^i) ) \big)
\end{equation}

\noindent where $P_{\mathrm{token}}$ denotes an MLP applied along the token dimension and consists of two linear layers with GeLU \cite{hendrycks2016gaussian} activation function inserted between them, $R_1$ and $R_2$ represent transposing the token and channel dimensions. The token mixture output $\boldsymbol{V}_E^i$ is obtained via a residual connection:
\begin{equation}
\boldsymbol{V}_E^i = \boldsymbol{V}_o^i + \boldsymbol{O}_{\mathrm{TM}}^i
\end{equation}

Next, the channel mixture operates along the channel dimension of each token to adaptively integrate cross-channel information:
\begin{equation}
\boldsymbol{O}_{\mathrm{CM}}^i = P_{\mathrm{channel}}(\boldsymbol{V}_E^i)
\end{equation}
where $P_{\mathrm{channel}}$ first increases channels of $\boldsymbol{V}_E^i$ and then projects it back to $D$ channels. The final reconstructed feature matrix $\boldsymbol{O}^i\in\mathbb{R}^{B\times(M_i+N)\times D}$ is obtained via a residual connection:
\begin{equation}
\boldsymbol{O}^i = \boldsymbol{V}_E^i + \boldsymbol{O}_{\mathrm{CM}}^i
\end{equation}

\subsubsection{Training}

Our model is trained using a weighted combination of Dice Loss and Weighted Cross-Entropy Loss:

\begin{equation}
\mathcal{L}_{\mathrm{total}} = (1 - \lambda) \, \mathcal{L}_{\mathrm{CE}} + \lambda \, \mathcal{L}_{\mathrm{Dice}}, \quad \lambda \in [0,1]
\end{equation}

\noindent where $\lambda$ is a hyperparameter controlling the relative contribution of Dice Loss and Cross-Entropy Loss during training. In our experiments, $\lambda$ is set to 0.1. 

\subsubsection{Inference}
During the inference stage, all input images are uniformly resized to a resolution of $480 \times 480$ and fed into the model for inference.

\section{Experiments}
\subsection{Implementation Details}
\subsubsection{Experiment Settings}
The adopted Swin Transformer is pre-trained on ImageNet-22K\cite{russakovsky2015imagenet}. Our proposed ICIPNet is optimized using AdamW\cite{loshchilov2018decoupled} with 40 epochs in total. The weight decay and learning rate are set to 0.01 and $3\times10^{-5}$. The number of channels $D$ is set to 768. We apply a polynomial decay learning rate scheduler to gradually reduce the learning rate during training. All experiments are conducted on a single NVIDIA GeForce RTX~3090 GPU with a batch size $B$ of~4.
\begin{figure*}[!t]
\vspace{-0.5mm}    
\centering
\includegraphics[width=\linewidth]{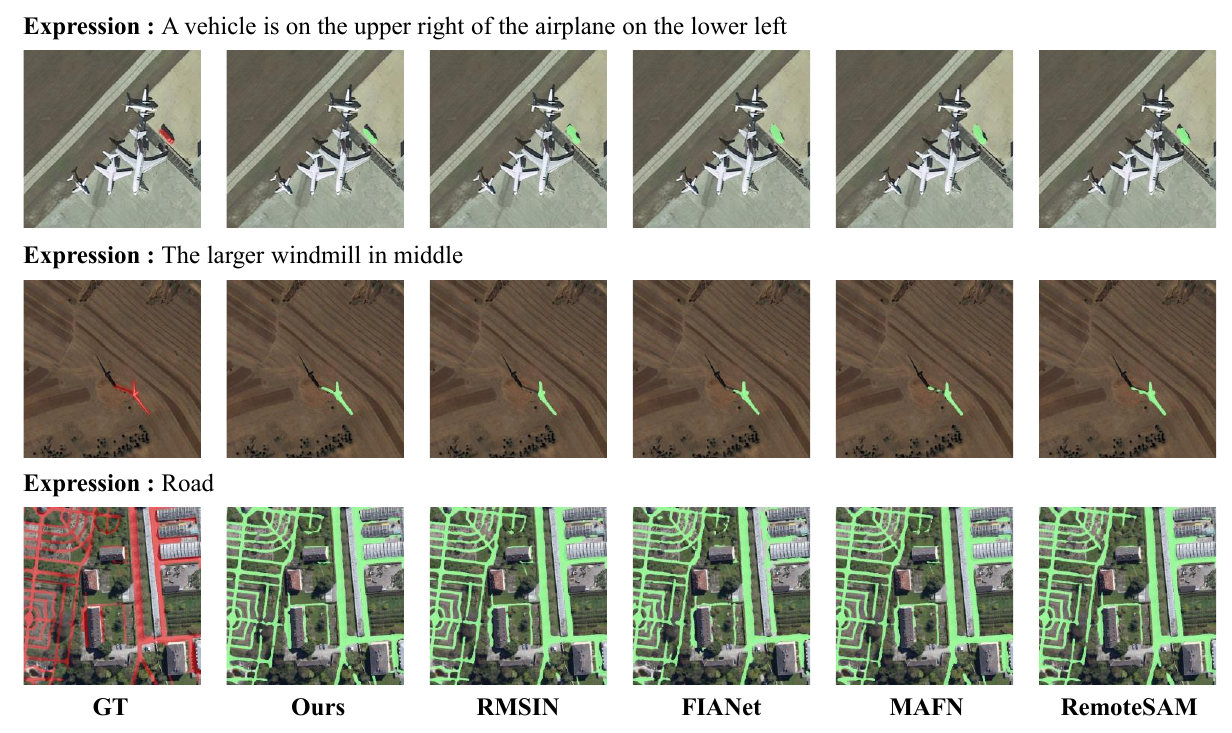}   
\vspace{-6mm}
\caption{Qualitative results of the proposed model and other existing models across different scenarios.}
\vspace{-4mm}
\label{figure_4_1}
\end{figure*}
\subsubsection{Metrics}
We evaluate the model using Overall Intersection-over-Union (oIoU), Mean Intersection-over-Union (mIoU), and Precision@X (P@X), where P@X is computed under IoU thresholds ranging from 0.5 to 0.9.
\subsection{Comparison with State-of-the-Art RIS Methods}
We conduct experiments of the proposed method and several recent RRSIS models on the validation and test sets of the RefSegRS and RRSIS-D datasets. To ensure fair comparison, we re-implemented the recent RRSIS works above, including RMSIN\cite{liu2024rotated}, 
FIANet\cite{lei2024exploring},
RemoteSAM\cite{yao2025remotesam},
and MAFN\cite{shi2025multimodal}, with the number of visual prompts $M_i$ at the $i$-th ST layer equaling to 8.

Table~\ref{tab:refsegrs} reports performances of different works on the RefSegRS dataset. 
The evaluation results on both the validation and test sets indicate that our proposed ICIPNet outperforms other existing works on many results. 
On the validation set, the model surpasses the runner-up model RemoteSAM by \textbf{0.94\%} and \textbf{1.09\%} in terms of mIoU and oIoU, respectively. 
On the test set, our proposed model follows the same trend that exceeds the runner-up model RemoteSAM on mIoU and FIANet on oIoU by \textbf{3.18\%} and \textbf{1.61\%}, respectively. 
These results further demonstrate the effectiveness of our method in cross-modal feature alignment and semantic modeling.

Table~\ref{tab:rrsis-d} reports the performances of different models on the RRSIS-D dataset. 
Results on both the validation and test sets indicate that our method outperforms other models on most results. 
On the validation set, our method improves over the runner-up models RemoteSAM on mIoU and MAFN on oIoU by \textbf{1.08\%} and \textbf{0.46\%}, respectively. 
On the test set, our model exceeds the runner-up model RemoteSAM in both mIoU and oIoU by \textbf{0.62\%} and \textbf{0.46\%}, respectively. 
These results further demonstrate the effectiveness of our method for fine-grained referring remote sensing image segmentation tasks.

\label{sec:exp}

\subsection{Ablation study}
To systematically evaluate the effectiveness of the pivotal components in ICIPNet, we conduct a series of ablation studies on the RefSegRS validation set. The experiments implement RMSIN as the base model, and the ICIP and BIF modules are introduced individually for comparative evaluation. The number of visual prompts in the $i$-th ST encoder layer $M_i$ is set to 5 in the ICIP module.



\begin{table}[!t]
\centering
\caption{Ablation study on ICIP and BIF modules. \ding{51} and \ding{55} indicate whether a module is included or not. The best results are in bold.}
\label{tab:AblationStudy}

\resizebox{\linewidth}{!}{
\begin{tabular}{c c | c c c c c c c}
\toprule
ICIP & BIF & P@0.5 & P@0.6 & P@0.7 & P@0.8 & P@0.9 & mIoU & oIoU \\
\midrule
\ding{55} & \ding{55} & 93.04 & 88.40 & 77.26 & 33.87 & 7.42 & 74.06 & 81.58 \\
\ding{51} & \ding{55} & 94.90 & 92.34 & 82.60 & 40.84 & 9.28 & 76.27 & 82.97 \\
\ding{55} & \ding{51} & 95.59 & 91.42 & 84.92 & 49.19 & \textbf{11.37} & 77.62 & 83.53 \\
\ding{51} & \ding{51} & \textbf{96.22} & \textbf{93.85} & \textbf{87.30} & \textbf{51.64} & 9.49 & \textbf{78.21} & \textbf{83.77} \\
\bottomrule
\end{tabular}
}
\vspace{-3mm}
\end{table}

The quantitative results for each model are reported in Table~\ref{tab:AblationStudy}. As shown in Table~\ref{tab:AblationStudy}, by adding the ICIP module alone, the mIoU and oIoU are increased by \textbf{2.21\%} and \textbf{1.39\%}, respectively, which highlights the model’s enhanced ability to capture key instance-level features. With the addition of the BIF module, mIoU and oIoU are further improved by \textbf{3.56\%} and \textbf{1.95\%}, indicating that this module can optimize feature distribution and enhance model's perceptual capability. When the ICIP module and the BIF module are both integrated, the model achieves the best performance across most metrics, where the mIoU is rised to \textbf{4.15\%} and the oIoU increases by \textbf{2.19\%}, demonstrating significant synergistic effects of our proposed two modules.

\begin{table}[!t]
\centering
\caption{Ablation study of different $M_i$ in the ICIP module on the RefSegRS test set.}
\label{tab:icip_ablation}
\resizebox{\linewidth}{!}{
\begin{tabular}{l|c|c|c|c|c|c|c}
\toprule
Options & P@0.5 & P@0.6 & P@0.7 & P@0.8 & P@0.9 & mIoU & oIoU \\
\midrule
Base Model & 75.45 & 63.73 & 38.14 & 14.14 & 2.97 & 60.23 & 74.64 \\
+ICIP ($M_i$=3) & 78.65 & 67.64 & 45.18 & 18.16 & 3.52 & 62.72 & 75.60 \\
+ICIP ($M_i$=5) & 79.97 & 69.40 & 46.01 & 18.44 & 3.52 & 63.40 & 75.85 \\
+ICIP ($M_i$=8) & 80.02 & 68.13 & 43.31 & 17.94 & 3.96 & 63.07 & 75.77 \\
\bottomrule
\end{tabular}
}
\vspace{-3mm}
\end{table}

\subsection{Visual Analysis}
As shown in \hyperref[figure_4_1]{Fig.~3}, our model accurately identifies targets across different scenarios, whereas baseline models exhibit broken prediction masks or noticeable deviations.

\subsection{Effect of the number of Learnable Instance Prompts in ICIP}

To investigate the effect of different numbers of visual prompts $M_i$ in the ICIP module, we conduct a series of experiments on the RefSegRS test set. As shown in Table~\ref{tab:icip_ablation}, the model consistently outperforms the RMSIN\cite{liu2024rotated} baseline across all configurations, indicating that the proposed method can effectively enhance visual–text feature modeling. 

\subsection{Performance on Each Category of the RefSegRS Dataset}


Table~\ref{tab:table} presents fine-grained mIoU results, showing performance variations across ground objects; for instance, segmenting “road marking” is relatively challenging, whereas segmenting “building” is comparatively easier. Our ICIPNet achieves substantially higher average and per-category mIoU than RMSIN, MAFN, FIANet, and RemoteSAM, demonstrating its effectiveness in handling diverse targets.
\begin{table}[!t]
\label{tab:table}
\centering
\caption{Results on each category of the RefSegRS dataset. The best results are shown in bold.}
\label{tab:method_comparison}
\renewcommand{\arraystretch}{1.3}
\resizebox{\columnwidth}{!}{%
\begin{tabular}{l|ccccc}
\hline
Category & RMSIN & MAFN & FIANet & RemoteSAM & Ours \\
\hline
impervious surface & 82.29 & 81.21 & 82.25 & 79.40 & \textbf{82.85} \\
low vegetation & 53.66 & 65.23 & 60.18 & 57.67 & \textbf{70.31} \\
road marking & 23.28 & 24.07 & \textbf{25.12} & 23.51 & 24.86 \\
paved road & 74.94 & 74.22 & 77.91 & 76.00 & \textbf{79.00} \\
building & 89.99 & 90.63 & 91.99 & 89.88 & \textbf{92.32} \\
sidewalk & 67.85 & 67.27 & 71.60 & 67.05 & \textbf{72.94} \\
vehicle & 76.06 & 77.62 & 78.37 & 76.21 & \textbf{78.62} \\
trailer & \textbf{80.18} & 80.02 & 76.41 & 76.11 & 77.83 \\
bikeway & 63.67 & 65.15 & 65.59 & 62.06 & \textbf{67.45} \\
truck & 73.64 & 76.63 & \textbf{85.43} & 82.78 & 81.80 \\
tree & 63.17 & 62.77 & 67.83 & 65.77 & \textbf{69.07} \\
car & 75.99 & 76.73 & 77.21 & 74.67 & \textbf{78.07} \\
van & 73.86 & 74.54 & 75.59 & 68.79 & \textbf{75.89} \\
\hline
average & 69.12 & 70.47 & 71.96 & 69.22 & \textbf{73.15} \\
\hline
\end{tabular}%
}
\vspace{-3mm}
\end{table}

\section{CONCLUSION}
In this paper, we propose ICIPNet for referring remote sensing image segmentation. Specifically, the Image-Conditioned Instance Prompt (ICIP) module generates adaptive visual prompts, allowing the model to acquire dynamic vision-language features of target instances, thereby effectively alleviating performance bottlenecks in cross-modal feature fusion. In addition, we design a Bilateral Information Fusion (BIF) strategy to optimize and reconstruct the cross-modal feature distribution. Our approach has been validated on two publicly available referring remote sensing datasets, RefSegRS and RRSIS-D. Future research may focus on developing more efficient visual prompt tuning methods within the embodied remote sensing perception, enabling models to acquire information through environment interaction and task-driven active exploration, thereby further enhancing their applicability in real-world scenarios.

\section*{Acknowledgment}
This work was supported by the National Natural Science Foundation of China (No. 62502387), the China Postdoctoral Science Foundation (No. BX20250486 \& No. 2025M784419), the Natural Science Basic Research Program of Shaanxi (No. 2025JC-YBQN-861), the Guangdong Basic and Applied Basic Research Foundation (No. 2025A1515011465) and the Postdoctoral Science Foundation of Shaanxi Province (No. 2025BSHSDZZ105).

\bibliographystyle{IEEEbib}
\bibliography{icme2026references}
\end{document}